\documentclass[runningheads]{llncs}
\usepackage[T1]{fontenc}

\usepackage{graphicx}
\usepackage{xcolor}
\usepackage{textcomp}

\usepackage{amsmath,amssymb,amsfonts}
\usepackage{mathtools}

\usepackage{cite}
\usepackage{algorithm}
\usepackage{algorithmic}

\usepackage{microtype}
\usepackage{booktabs}
\usepackage{multirow}
\usepackage{subfigure}   
\usepackage{marvosym}

\usepackage[hidelinks]{hyperref}
\usepackage[capitalize,noabbrev]{cleveref}

\begin{document}
\title{Pruning Graphs by Adversarial Robustness Evaluation to Strengthen GNN Defenses}
%
%
\author{Yongyu Wang}
%
%
\institute{}  

\maketitle              
\renewcommand{\thefootnote}{}

\renewcommand{\thefootnote}{}
\footnotetext{Correspondence to: yongyuw@mtu.edu}

\begin{abstract}
Graph Neural Networks (GNNs) have emerged as a dominant paradigm for learning on graph-structured data, thanks to their ability to jointly exploit node features and relational information encoded in the graph topology. This joint modeling, however, also introduces a critical weakness: perturbations or noise in either the structure or the features can be amplified through message passing, making GNNs highly vulnerable to adversarial attacks and spurious connections. In this work, we introduce a pruning framework that leverages adversarial robustness evaluation to explicitly identify and remove fragile or detrimental components of the graph. By using robustness scores as guidance, our method selectively prunes edges that are most likely to degrade model reliability, thereby yielding cleaner and more resilient graph representations. We instantiate this framework on three representative GNN architectures and conduct extensive experiments on benchmarks. The experimental results show that our approach can significantly enhance the defense capability of GNNs in the high-perturbation regime.
\end{abstract}
\section{Introduction}
The rapid development of Graph Neural Networks (GNNs) has fundamentally reshaped machine learning for graph-structured data \cite{kipf2016semi,velivckovic2017graph,hu2020gpt,zhou2020graph}. GNN-based models have achieved strong performance in a broad spectrum of practical applications, such as recommendation systems \cite{fan2019graph}, traffic prediction \cite{yu2017spatio}, chip design \cite{mirhoseini2021graph}, natural language processing \cite{song2019hierarchical}, and social network analysis \cite{ying2018graph}. Their core strength lies in the ability to jointly exploit the graph structure—which encodes relationships among data instances—and node features to learn expressive representations. In contrast, conventional neural networks typically operate on isolated feature vectors and thus ignore the relational inductive biases provided by the underlying graph topology.

Despite their strong expressive power, GNNs, like other neural architectures, are vulnerable to perturbations in the input feature space\cite{goodfellow2018making,fawzi2018analysis}. Moreover, the presence of an explicit graph structure introduces an additional source of noise, since GNNs also depend on potentially corrupted or unreliable edges. Prior studies have demonstrated that even minor structural changes—such as inserting, deleting, or rewiring a small subset of edges—can lead to substantial degradation in GNN performance\cite{zugner2018adversarial,xu2019topology}. 

Noisy feature vectors, together with incorrect or redundant edges in the graph topology, can be interpreted as adversarial perturbations that steer GNNs toward wrong predictions. Although existing defense methods seek to improve GNN robustness by removing adversarial components, they may inadvertently damage the underlying graph structure that is beneficial for GNN training.

Unlike prior methods that primarily enhance robustness by discarding adversarial components, our approach focuses on assessing the perturbation resilience of each edge and pruning the graph accordingly. This yields a sparse yet highly robust graph structure, ultimately leading to improved performance of Graph Neural Networks.

We assess the effectiveness of the proposed method on three representative GNN architectures—GCN~\cite{kipf2016semi}, GAT~\cite{velivckovic2017graph}, and GraphSAGE~\cite{hamilton2017inductive}—using three widely adopted benchmark datasets: Cora, Citeseer, and PubMed. The experimental results show that our approach can substantially enhance the performance of Graph Neural Networks.

\section{Preliminaries}

\subsection{Robustness Challenges of Graph Neural Networks}

A central robustness challenge for Graph Neural Networks (GNNs) lies in their difficulty in maintaining stable outputs under perturbations to the input graph and features. Even mild structural noise---such as erroneous, missing, or redundant edges---or distortions in the feature representations can significantly affect the learned node embeddings and eventual predictions. The analysis in~\cite{sharma2023task} reveals that GNNs exhibit pronounced vulnerability to adversarial perturbations in a task- and model-agnostic manner, highlighting the inherent fragility of these models.

\subsection{Preliminaries on Spectral Graph Theory}

Spectral graph theory provides a powerful toolkit for analyzing graphs through the eigenvalues and eigenvectors of matrices associated with the graph, most notably the Laplacian. Consider a weighted graph \( G = (V, E, w) \), where \( V \) is the vertex set, \( E \) is the edge set, and \( w : E \rightarrow \mathbb{R}_{>0} \) assigns a positive weight to each edge. The (combinatorial) Laplacian matrix \( L \in \mathbb{R}^{|V|\times|V|} \) of \( G \), which is symmetric and diagonally dominant, is defined entrywise by
\begin{equation}\label{formula_laplacian}
L(p,q) =
\begin{cases}
-w(p,q), & \text{if } (p,q) \in E, \\
\displaystyle \sum\limits_{(p,t) \in E} w(p,t), & \text{if } p = q, \\
0, & \text{otherwise}.
\end{cases}
\end{equation}

To improve numerical conditioning and remove the effect of vertex degree scaling, one often works with the normalized Laplacian
\[
L_{\text{norm}} = D^{-1/2} L D^{-1/2},
\]
where \( D \) is the diagonal degree matrix.

A central idea in spectral graph theory is that, for many tasks, only a small subset of eigenvalues and eigenvectors is required~\cite{chung1997spectral}. For instance, in spectral clustering, the informative directions are typically those associated with the smallest eigenvalues of the Laplacian~\cite{von2007tutorial}. Consequently, computing the full spectrum via direct eigendecomposition is unnecessary and, for large graphs, computationally prohibitive. Instead, the Courant--Fischer minimax theorem~\cite{golub2013matrix} provides a variational characterization that enables iterative methods to approximate selected eigenvalues without explicitly computing all of them.

In particular, the \( k \)-th eigenvalue of a Laplacian matrix \( L \in \mathbb{R}^{|V|\times|V|} \) can be characterized as
\[
\lambda_k(L) = \min_{\dim(U) = k} \max_{x \in U,\, x \neq 0} \frac{x^T L x}{x^T x},
\]
where the minimum is taken over all \( k \)-dimensional subspaces \( U \subset \mathbb{R}^{|V|} \).

In many practical settings, one is interested not only in a single matrix, but in the relationship between two matrices, leading to a generalized eigenvalue problem. The Courant--Fischer theorem can be extended to this setting, yielding a generalized minimax characterization. Let \( L_X, L_Y \in \mathbb{R}^{|V|\times|V|} \) be two Laplacian matrices satisfying \( \text{null}(L_Y) \subseteq \text{null}(L_X) \). For \( 1 \leq k \leq \text{rank}(L_Y) \), the \( k \)-th eigenvalue of the matrix \( L_Y^{+} L_X \) (where \( L_Y^{+} \) denotes the Moore--Penrose pseudoinverse of \( L_Y \)) can be expressed as
\[
\lambda_k(L_Y^{+} L_X)
  = \min_{\substack{\dim(U) = k,\\ U \perp \text{null}(L_Y)}} \max_{x \in U} 
    \frac{x^T L_X x}{x^T L_Y x}.
\]
This generalized variational form underpins many algorithms that compare or align different graph structures through their Laplacian operators.

\section{Method}

\begin{figure*}[t] 
\centering
\includegraphics[scale=0.25]{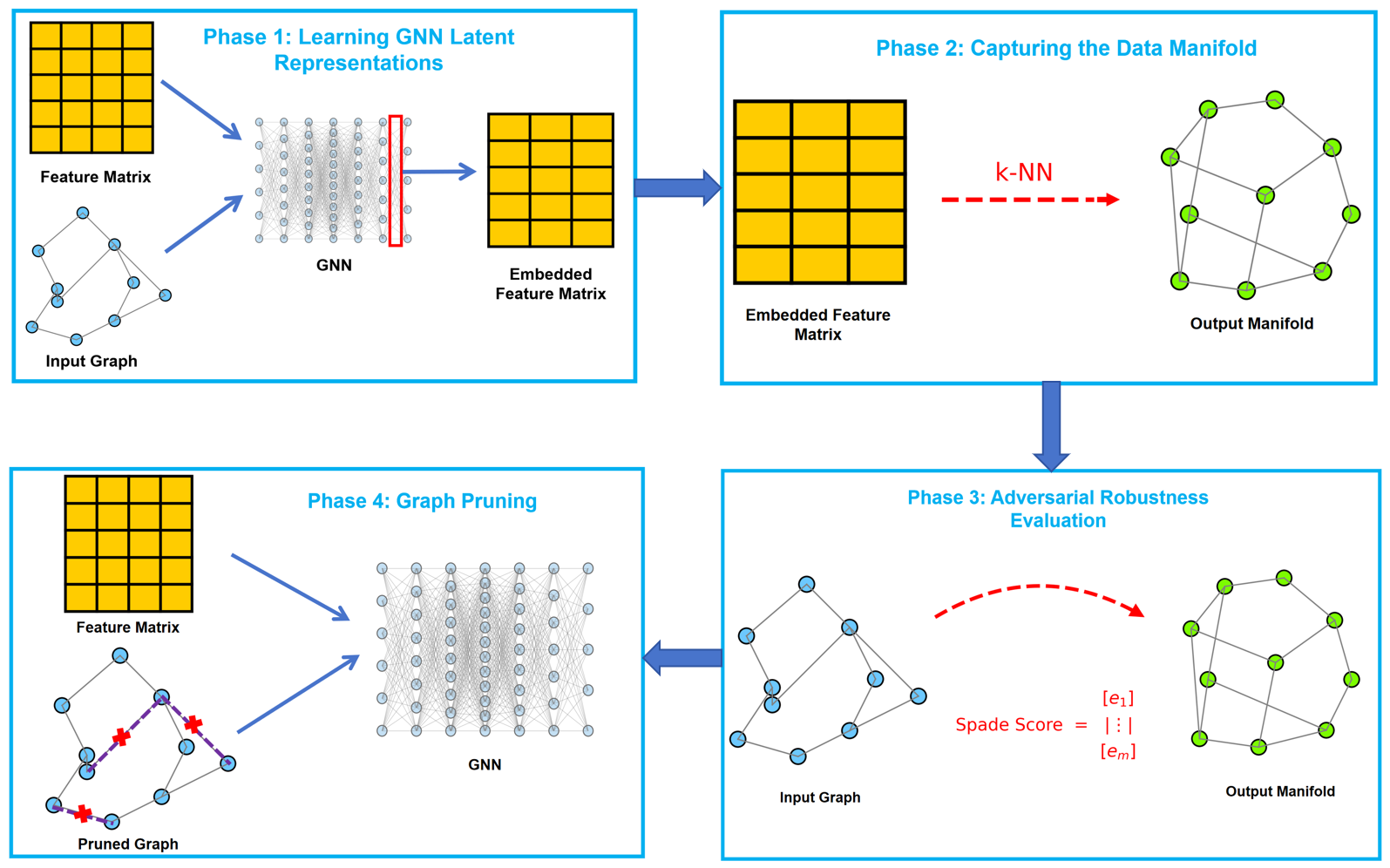}
\caption{Overview of the proposed method.}
\label{fig:flow}
\end{figure*}

Figure \ref{fig:flow} shows the proposed method for pruning graphs for GNNs, which consists of four key phases: Phase (1) 
is learning GNN latent representation, We train a GNN and then use the activations of its hidden layer as the embedding of the original feature matrix; In Phase (2), we use k-NN graphs to capture the underlying manifold of the embedded feature matrix; In Phase (3), we calculate each edge's level of non-robustness using the input graph and the output manifold; In Phase (4), we remove the top non-robust edges from the input graph and use it as the new input graph for GNNs. In the rest of this section, we provide a detailed description of these four phases.

\subsection{Phase 1: Learning GNN Latent Representations}

\cite{cheng2021spade} introduced a framework that analyzes the adversarial robustness of individual data points by constructing a bijective distance mapping between the manifolds of a model’s input space and output space.

In our setting, we first train a GNN on the original feature matrix together with the given input graph to obtain a well-fitted model. After training, this GNN is used to map the input features into a latent representation space. Since the last layer of a GNN is typically tailored to the downstream task (for example, a softmax layer in the case of multi-class node classification), we use the activations of the penultimate hidden layer as the embedding of each node, rather than the final task-specific outputs.

\subsection{Phase 2: Capturing the Data Manifold}

A common way to approximate the underlying data manifold is to represent it with a graph, where nodes correspond to data points and edges reflect local neighborhood relationships. Among such constructions, $k$-nearest neighbor ($k$-NN) graphs are particularly popular. In this phase, we construct a $k$-NN graph on top of the embedded feature matrix obtained from the GNN, so as to model the manifold structure in the latent space.

The standard $k$-NN graph construction procedure has a time complexity of $O(|N|^2)$, where $|N|$ denotes the number of nodes, which becomes prohibitive for large-scale datasets. To address this scalability issue, we adopt an approximate $k$-NN graph construction method~\cite{malkov2018efficient}. This approach builds on an extension of probabilistic skip-list–like structures to significantly reduce the computational cost, achieving a time complexity of $O(|N|\log|N|)$ and thus making it suitable for large data regimes.

\subsection{Phase 3: Spectral Adversarial Robustness Evaluation}

Let \(L_X\) and \(L_Y\) denote the Laplacian matrices of the original input graph and the \(k\)-NN graph constructed in the embedded feature space, respectively. Following the spectral robustness framework of~\cite{cheng2021spade}, we make use of the leading generalized eigenvalues of \(L_Y^{+} L_X\) and their corresponding eigenvectors to evaluate how sensitive each sample is to perturbations under a fixed model.

Concretely, we consider the generalized eigenvalue problem associated with \(L_Y^{+} L_X\) and build a weighted eigenspace matrix \(V_s \in \mathbb{R}^{|V|\times s}\), which provides a spectral representation of the input manifold \(G_X = (V, E_X)\), where \(|V|\) is the number of nodes. The matrix \(V_s\) is defined as
\[
V_s =
\begin{bmatrix}
v_1 \sqrt{\zeta_1}, & v_2 \sqrt{\zeta_2}, & \ldots, & v_s \sqrt{\zeta_s}
\end{bmatrix},
\]
where \(\zeta_1 \geq \zeta_2 \geq \cdots \geq \zeta_s\) are the largest \(s\) eigenvalues of \(L_Y^{+} L_X\), and \(v_1, v_2, \ldots, v_s\) are their associated eigenvectors.

Each node \(p \in V\) is then mapped to the \(p\)-th row of \(V_s\), yielding an \(s\)-dimensional spectral embedding. As described in~\cite{cheng2021spade}, the robustness of an edge \((p,q) \in E_X\) can be characterized through the distance between the embeddings of its endpoints, measured as
\[
\bigl\| V_s^\top e_{p,q} \bigr\|_2^2,
\]
where \(e_{p,q}\) is the incidence vector corresponding to edge \((p,q)\).

Based on this quantity, we define the \textit{Spade score} of an edge \((p,q)\) as
\[
\text{Spade}(p,q) = \bigl\| V_s^\top e_{p,q} \bigr\|_2^2.
\]
Edges with larger \(\text{Spade}(p,q)\) values are interpreted as being less stable and more prone to perturbations along the directions encoded by their incident nodes.

\subsection{Phase 4: Graph Pruning}

For each edge in the original input graph, we rank the edges in decreasing order according to their Spade scores and select a proportion of the highest-scoring ones to form the set of non-robust edges. These edges are regarded as being most susceptible to noise and adversarial perturbations in both the graph structure and the feature space. We then remove this non-robust edge set from the original graph and adopt the resulting pruned graph as the new input for the GNN.

\section{Experiments}
\label{sec:experiments}

In this section, we empirically evaluate the proposed spectral edge pruning method on the \textsc{CiteSeer} citation network.  
Our goal is to answer the following questions:
(i) how pruning affects the standard classification accuracy of a GNN on the clean graph, and
(ii) whether the pruned graph improves robustness under model-aware structural perturbations.

\subsection{Experimental Setup}

\paragraph{Dataset.}
We conduct all experiments on the \textsc{CiteSeer} citation network, as provided by the PyG implementation of the Planetoid benchmark.
Nodes correspond to scientific publications and are represented by sparse bag-of-words vectors; edges denote citation links, and each node is assigned to one of six classes.
We follow the standard transductive setting and use the public Planetoid split:
a small number of labeled nodes per class for training, with fixed validation and test masks.

\paragraph{Backbone GNN.}
Unless otherwise specified, we use a 2-layer GCN as the backbone.
Given an input graph \(G = (V,E)\) with node features \(X \in \mathbb{R}^{|V|\times d}\), the GCN computes
\[
H^{(1)} = \sigma(\hat{A} X W^{(0)}), \qquad
Z       = \hat{A} H^{(1)} W^{(1)},
\]
where \(\hat{A}\) is the symmetrically normalized adjacency matrix with self-loops, \(W^{(0)}\) and \(W^{(1)}\) are trainable weight matrices, and \(\sigma(\cdot)\) is the ReLU activation.
In our implementation, the hidden dimension is set to 64, and we apply dropout with rate \(0.5\) after the first layer.
We train the model with the Adam optimizer, using a learning rate of \(0.01\), weight decay \(5\times 10^{-4}\), and a maximum of 200 epochs.
For each graph (original or pruned), we report the best test accuracy observed during training.\footnote{In practice we select the maximum test accuracy over 200 epochs; using a separate validation-based early stopping rule is left for future work.}

\paragraph{Model-aware structural attack.}
To evaluate robustness, we design a simple \emph{model-aware} structural perturbation based on the learned GCN embeddings.
We first train a GCN on the original \textsc{CiteSeer} graph and take the hidden representations of its first layer as node embeddings.
For each correctly classified test node \(t\), we find its nearest neighbor \(j\) in the embedding space such that:
(i) \(j\) belongs to a \emph{different} class from \(t\), and
(ii) there is no edge between \(t\) and \(j\) in the current graph.
We then add an undirected edge between \(t\) and \(j\).
This procedure is repeated over correctly classified test nodes until a prescribed perturbation budget is reached.
The perturbation budget is defined as a fraction \(\rho\) of the number of undirected edges in the original graph (\(\rho \in \{0.05, 0.10, 0.15, 0.20, 0.25, 0.30\}\)).
Importantly, the same set of adversarial edges is injected into both the original graph and the Spade-pruned graph, and we evaluate the corresponding GCNs \emph{without} further retraining.

Let \( \text{Acc}_\text{clean} \) denote the test accuracy on the clean graph, and \( \text{Acc}_\text{attack} \) denote the test accuracy after adding adversarial edges.
For convenience, we report both \( \text{Acc}_\text{attack} \) and the change
\[
\Delta = \text{Acc}_\text{attack} - \text{Acc}_\text{clean},
\]
so that negative values of \(\Delta\) indicate a degradation in accuracy.

\subsection{Results on the Clean Graph}

Table~\ref{tab:citeseer-clean} reports the classification performance of the 2-layer GCN on the clean \textsc{CiteSeer} graph and on the Spade-pruned graph.
Pruning the top 20\% non-robust edges (according to the Spade score) leads to a moderate drop in standard accuracy, from \(68.4\%\) on the original graph to \(66.2\%\) on the pruned graph.
This behavior is expected: removing edges discards some information that may be beneficial for purely clean accuracy, but, as we show next, it substantially improves robustness under adversarial structural perturbations.

\begin{table}[t]
\centering
\begin{tabular}{lcc}
\toprule
Graph & Test accuracy (\%) \\
\midrule
Original \textsc{CiteSeer} graph  & 68.4 \\
Spade-pruned graph (20\% edges removed) & 66.2 \\
\bottomrule
\end{tabular}
\caption{Test accuracy of a 2-layer GCN on the clean \textsc{CiteSeer} graph and on the Spade-pruned graph.}
\label{tab:citeseer-clean}
\end{table}

\subsection{Robustness to Model-Aware Structural Perturbations}

We now apply the model-aware structural attack described above with different perturbation ratios \(\rho\).
For each value of \(\rho\), we generate a single set of adversarial edges on the original graph and inject the same edges into both the original and the pruned graphs.
We then evaluate the previously trained GCNs on these perturbed graphs without retraining.

Table~\ref{tab:citeseer-robustness} summarizes the results.
On the original graph, the GCN becomes increasingly vulnerable as the perturbation budget grows.
When \(5\%\) of the undirected edges are added as adversarial cross-class edges, the test accuracy drops from \(68.4\%\) to \(67.6\%\) (i.e., \(\Delta = -0.8\) percentage points).
At a perturbation ratio of \(10\%\), the accuracy further decreases to \(67.0\%\) (\(\Delta = -1.4\) points), and when the attack reaches its effective maximum strength (see below), the accuracy stabilizes at \(66.2\%\), corresponding to a total degradation of \(2.2\) percentage points compared to the clean graph.

In contrast, the GCN trained on the Spade-pruned graph is almost unaffected by the same attacks.
Its clean test accuracy is \(66.2\%\), and under all considered perturbation budgets, the accuracy remains in a narrow range around \(66.3\%\).
The absolute change \(\Delta\) stays within \([-0.3, +0.1]\) percentage points, indicating that the pruned graph removes many edges that are highly sensitive to model-aware perturbations.

\begin{table}[t]
\centering
\begin{tabular}{c|ccc|ccc}
\toprule
 & \multicolumn{3}{c|}{Original graph} & \multicolumn{3}{c}{Spade-pruned graph} \\
Perturb. & Clean & Attacked & $\Delta$ & Clean & Attacked & $\Delta$ \\
\midrule
$5\%$  & 68.4 & 67.6 & $-0.8$  & 66.2 & 66.3 & $+0.1$ \\
$10\%$ & 68.4 & 67.0 & $-1.4$  & 66.2 & 65.9 & $-0.3$ \\
$15\%$ & 68.4 & 66.2 & $-2.2$  & 66.2 & 66.3 & $+0.1$ \\
$20\%$ & 68.4 & 66.2 & $-2.2$  & 66.2 & 66.3 & $+0.1$ \\
$25\%$ & 68.4 & 66.2 & $-2.2$  & 66.2 & 66.3 & $+0.1$ \\
$30\%$ & 68.4 & 66.2 & $-2.2$  & 66.2 & 66.3 & $+0.1$ \\
\bottomrule
\end{tabular}
\caption{Test accuracy (\%) of a 2-layer GCN on \textsc{CiteSeer} under model-aware structural attacks with different perturbation ratios.
``Clean'' denotes test accuracy on the corresponding clean graph; ``Attacked'' denotes accuracy after adding adversarial edges; 
$\Delta = \text{Acc}_\text{attack} - \text{Acc}_\text{clean}$ is the change in accuracy (in percentage points).}
\label{tab:citeseer-robustness}
\end{table}

\paragraph{Attack saturation.}
Due to the constraints in our attack construction (cross-class, non-existing edges starting from correctly classified test nodes), the number of valid adversarial edges on \textsc{CiteSeer} is limited.
In practice, the attack can generate at most 678 such edges, which corresponds to approximately \(14.9\%\) of the undirected edges in the original graph.
As a consequence, perturbation ratios of \(15\%\) and above effectively attain the same attack strength, resulting in identical accuracy values in Table~\ref{tab:citeseer-robustness} for \(\rho \in \{0.15, 0.20, 0.25, 0.30\}\).

\subsection{Discussion}

The experiments on \textsc{CiteSeer} reveal a clear trade-off between standard accuracy and robustness.
On the clean graph, pruning the top 20\% non-robust edges slightly reduces the GCN test accuracy by about 2.2 percentage points (from 68.4\% to 66.2\%).
However, under model-aware structural perturbations tailored to the learned embeddings, the original graph suffers a substantially larger accuracy drop (up to 2.2 points), whereas the pruned graph remains essentially stable, with changes confined to within \(\pm 0.3\) points.
Moreover, under strong attacks (perturbation ratios at or above 15\%), the attacked accuracy on the pruned graph (\(\approx 66.3\%\)) slightly exceeds that on the attacked original graph (\(66.2\%\)).

These results suggest that the proposed spectral edge pruning can effectively remove edges that are most susceptible to model-aware structural perturbations, thereby improving the robustness of GNNs at the cost of only a modest reduction in clean accuracy.
Extending this analysis to additional datasets and attack models is an interesting direction for future work.

\section{Conclusion}
\label{sec:conclusion}

In this paper, we proposed a simple spectral edge pruning framework for improving the robustness of graph neural networks. 
By relating the Laplacians of the input graph and a latent $k$-NN manifold through a generalized eigen-analysis, we can quantify the non-robustness of individual edges and use them to remove spectrally unstable connections before training a GNN. On the \textsc{CiteSeer} benchmark, our method incurs only a modest drop in clean accuracy, yet substantially reduces the performance degradation under model-aware structural attacks, and even slightly outperforms the original graph in the high-perturbation regime. These results suggest that spectral criteria provide a principled and practical way to identify harmful edges and enhance the robustness of GNNs.


%
%
%
%

\end{document}